# Enhancements in statistical spoken language translation by de-normalization of ASR results

Agnieszka Wołk[1]*, Krzysztof Wołk[2], Krzysztof Marasek[3]

[1] Department of Cybernetics, Military University of Technology, Kaliskiego 2, Warsaw. Poland
[2] Polish-Japanese Institute of Information Technology, Koszykowa 86, Warsaw, Poland.
[3] Polish-Japanese Institute of Information Technology, Koszykowa 86, Warsaw, Poland.

* Corresponding author. Tel.: +48 725891122 email: agnieszka.wolk@student.wat.edu.pl

**Abstract:** Spoken language translation (SLT) has become very important in an increasingly globalized world. Machine translation (MT) for automatic speech recognition (ASR) systems is a major challenge of great interest. This research investigates that automatic sentence segmentation of speech that is important for enriching speech recognition output and for aiding downstream language processing. This article focuses on the automatic sentence segmentation of speech and improving MT results. We explore the problem of identifying sentence boundaries in the transcriptions produced by automatic speech recognition systems in the Polish language. We also experiment with reverse normalization of the recognized speech samples.

## 1. Introduction

In natural spoken language there are many meaningless modal particles and dittographes. Furthermore, ASR often produces some recognition errors, and the ASR results have no punctuation or proper casing. Therefore, the translation would be rather poor if the ASR results are directly translated by MT systems. Automatic sentence segmentation of speech is important to make speech recognition (ASR) output more readable and easier for downstream language processing modules. Various techniques have been studied for automatic sentence boundary detection in speech, including hidden Markov models (HMMs), maximum entropy, neural networks, and Gaussian mixture models, utilizing both textual and prosodic information.

This paper addresses the problem of identifying sentence boundaries in the transcriptions produced by automatic speech recognition systems. These differ from the sorts of texts normally used in NLP in a number of ways: the text is generally in single case, unpunctuated, and may contain transcription errors. In addition, such text is usually normalized, what is unnatural to read. A lot of important linguistic information is lost, as well (e.g., for machine translation or text-to-speech). Table 1 compares a short text in the format that would be produced by an ASR system with a fully punctuated version that includes casing information and de-normalization1.

There are many possible situations in which an NLP system may be required to process ASR text. The most obvious examples are text to speech systems, machine translation or real-time speech-to-speech translation systems.Dictation software programs do not punctuate or capitalize their output. However, if this information could be added to ASR text, the results would be far more useful.

One of the most important pieces of information that is not available in ASR output is sentence boundary information. The knowledge of sentence boundaries is required by many NLP technologies. Parts of speech taggers typically require input in the format of a single sentence per line, and parsers generally aim to produce a tree spanning each sentence. Sentence boundary is very important in order not to lose the context or even the meaning of a sentence. Only the most trivial linguistic analysis can be carried out on text that is not split into sentences.

It is worth mentioning that not all transcribed speech can be sensibly divided into sentences. It has been argued by Gotoh and Renals (2000), that the main unit in spoken language is the phrase, rather than the sentence. However, there are situations in which it is appropriate to consider spoken language to be composed of sentences. The need for such tools is also well-known in other NLP branches. For example, the spoken portion of the British National Corpus (Burnard, 1995) contains 10 million words and was manually marked with sentence boundaries. A technology that identifies sentence boundaries could be used to speed up such processes2.

Table 1. Sample input and output.

| | |
|---|---|
| Before | ile kobiet mających czterdzieści cztery lata wygląda tak jak amerykańska gwiazda na okładce swej nowej płyty jennifer lopez nie bierze jednak udziału w konkursie piękności czy więc również mocno jak do rzeźbienia swego ciała przyłożyła się do stworzenia dobrych piosenek ostatnie lata nie były dobre dla latynoskiej piosenkarki między innymi rozwody i rozstania zdewastowały jej poczucie kobiecej wartości a klapy ostatnich albumów o mały włos nie zrujnowały piosenkarskiej kariery |
| After | Ile kobiet mających 44 lata wygląda tak, jak amerykańska gwiazda na okładce swej nowej płyty Jennifer Lopez? Nie bierze jednak udziału w konkursie piękności. Czy więc również mocno, jak do rzeźbienia swego ciała, przyłożyła się do stworzenia dobrych piosenek?. Ostatnie lata nie były dobre dla latynoskiej piosenkarki, m.in. rozwody i rozstania zdewastowały jej poczucie kobiecej wartości, a klapy ostatnich albumów o mały włos nie zrujnowały piosenkarskiej kariery. |

It is important to distinguish the problem just mentioned from another problem sometimes called "sentence splitting". This problem aims to identify sentence boundaries in standard text. However, since standard text includes punctuation, the problem is effectively reduced to deciding which of the symbols that potentially denotes sentence boundaries (., !, ?) actually do. This problem is not trivial since these punctuation symbols do not always occur at the end of sentences. For example, in the sentence "Mr. Doe teacher at M.I.T." only the final full stop denotes the end of a sentence. For the sake of clarity, we shall refer to the process of discovering sentence boundaries in standard punctuated text as "punctuation disambiguation" and that of finding them in unpunctuated ASR text as "sentence boundary detection"3.

Sentence boundary detection is not all that is needed. We cannot forget about restoring proper casing, distinguishing between normal sentences and questions, or exclamatory sentences. Restoring numbers, dates, separators, etc. is not a trivial task, so it is still important to also obtain linguistic information from other tools.

## 2. Difficulties with the Polish language

The Polish language is a particular challenge to such tools. It is a very complicated West-Slavic language with complex elements and grammatical rules. In addition, the Polish language has a large vocabulary due to many endings and prefixes changed by word declension. These characteristics have a significant effect on the requirements for the data and the data structure.

In addition, English is a position-sensitive language. The syntactic order (the order of words in a sentence) plays a very significant role, and the language has very limited inflection of words (e.g., due to the lack of declension endings). The word position in an English sentence is often the only indicator of the meaning. The sentence order follows the Subject-Verb-Object (SVO) schema, in which the subject phrase preceding the predicate4.

On the other hand, no specific word order is imposed in Polish, and the word order has little effect on the meaning of a sentence. The same thought can be expressed in several ways. For example, the sentence "I bought myself a new car." can be written in Polish as one of the following: "Kupiłem sobie nowy samochód"; Nowy samochód sobie kupiłem."; "Sobie kupiłem nowy samochód."; "Samochód nowy sobie kupiłem.". It must be noted that such differences exist in many language pairs and need to be addressed in some manner7.

Moreover, the Polish language requires many rules that are hard to define and implement. First, we need to recognize names and surnames. A surname can also be a name at the same time (e.g., "Tobiasz" can be recognized as either the name "Tobiasz Kowalski" or the surname "Michał Tobiasz". Let us analyze the surname "Biały." In this scenario, the word "biały" also represents a name for the color white, so it is also a casual word. The same problems occur when we deal with names.

Another problem with names and surnames are salutations (e.g.,"Pan"/"Sir") and initials before, after, or even inside a name (e.g.,"Mr. John Doe," Mr. J. Doe," "A.K. Doe," Mr. John W Doe," etc.). Those have to be capitalized, punctuated and distinguished from sentence boundaries as well.

It is also important to restore all possible abbreviations. That task is not as trivial as it may sound. In some cases, abbreviations are also casual words. In the sentence "W wieku 13 lat." the word "wieku" is just a noun, but if we consider the sentence "W 21 wieku.", the word "21 wieku" means the 21st century and should be treated as an abbreviation. In the other words, to accomplish many tasks correctly, it is important to know the context of a sentence. Abbreviations also include names of popular organizations or institutions.

Knowledge of the context is also important to distinguish normal sentences from questions or exclamations. Exclamations can also be a part of a quotation (e.g.,"Zabij go! - powiedziałem.") or a single word (e.g.,"Łap!," "Gol!," etc.). We cannot forget about rhetorical questions that are sometimes hard to distinguish. For example, the exclamation "Zwarowiałeś!" can function as the statement "Zwariowałeś." or the rhetorical question "Zwariowałeś?", depending on the context.

Probably one the most problematic aspects is the conversion of written numbers into their Arabic or Roman form. It is important to prepare a set of regular expressions that could distinguish IP address, enumeration, telephone number, natal number, date, time, percentages, fraction, degrees, currency, cardinality, and ordinal numerals. Each of these types has to be processed in a different way.

Because of the differences in meaning and usage, we can distinguish cardinal numbers by answering the question "how many?" (e.g.,"pięć" - five, "sto siedemnaście" - one hundred and seventeen); ordinal numbers, by "which in turn?" (e.g.,"piąty" - fifth, "sto siedemnasty" - one hundred and by); collective numbers, by "how many (in a group)?" ("pięcioro" - five, "sto siedemnaścioro" - one hundred and seventeen); fractions (e.g.,"pół" - a half, "jedna piąta" - one fifth); and indeterminate numbers by stating the number in a non-exact way (e.g.,"kilka" - several, "kilkaset" - several hundred).

Cardinal numbers are declined by their case and gender, e.g.,"trzej uczniowie," "trzech uczniów" (three pupils). Ordinal numbers are declined as adjectives: by case, number, and gender. Collective numbers are declined only by case and must appear together with nouns. Fractional numbers are declined only by case. Indeterminate numbers are declined by case and gender. We distinguish 7 cases, 2 numbers and 5 genders in Polish, which causes great diversity among the numbers. In some situations, Roman numerals are used instead of Arabic numbers. In most cases, those numbers are correlated with the recognition of ordinal numbers.

Lastly, our objectives include implementing rules for placing commas in proper places in accordance with statistical models enriched with the Polish grammar rules.

## 3. De-Normalization Method

In order to achieve this task, we implement a tool that works in a specialized pipeline of minor tools. We wanted it to be as easy for non-IT specialists as possible. In the first step, we used a conditional random fields framework5 for building probabilistic models to segment and label sequential data. Conditional random fields in this scenario offered several advantages over other methods such as hidden Markov models and stochastic grammars. CRFs include the ability to relax strong independence assumptions made in the other models. They also avoid a fundamental limitation of maximum entropy Markov models and other discriminative Markov models, which can be biased towards states with few successor states. A CRF differs from the HMM with respect to its training objective function (joint versus conditional likelihood) and its handling of dependent word features. Traditional HMM training does not maximize the posterior probabilities of the correct labels; whereas, the CRF directly estimates posterior boundary label probabilities P(E|O). The underlying n-gram sequence model of an HMM does not cope well with multiple representations of a word sequence, especially when the training set is small. On the other hand, the CRF model supports simultaneously correlated features, providing more flexibility to incorporate a variety of knowledge sources6. The CRFSuite is used for sentence boundary detection, as well as for comma separation detection7.

For tasks such as conversion of numbers or capitalization, we created specialized rules. A specific python tool was implemented for this job. It required us to prepare large databases that were obtained using web crawlers, manually cleaned, and annotated. It was necessary to annotate, for example, surnames that could be either casual words or names. The same work was required for enumeration, abbreviations, numerals, currency, etc. The surnames database consisted of 399,551 records, 2769 names, and 171 abbreviations. We also created a database of special cases for the Polish language. In the program, we assured the ease of extending rules and the databases to support future research.

## 4. De-Normalization Experiments

In our experiments, we focused as much as possible on natural human speech in real-life scenarios. In order to do that, we chose three different types of data to process. One of the first corpora, similar to human dialogues, that we used was the Open Subtitles corpora obtained from the OPUS project9. In addition, we obtained multi-subject TED Lectures from the IWSLT'13 campaign proceedings and from the European Parliament Proceedings, whose subject domain vocabulary is very limited.

The corpora were randomly divided into training, tuning, and testing data. For training and tuning, we selected 1000 sentences for each task. In order to maximize the relevance of evaluation and coverage of the entire corpora, our randomization scripts first divided the corpora into 250 equal segments and then randomly chose four sentences from each segment. The selected sentences were removed from the training data.

Table 2 represents both the data specification and results of our initial sentence boundary detection script. The table rows represent: the number of sentences in each corpora, the number of words, the number of unique words and their forms, the average sentence length (words), and the measured accuracy of the trained system.

Table 2. Results of the tool.

|  | TED | EuroParl | OpenSubtitles |
|---|---|---|---|
| Sentences | 185,637 | 629,558 | 9,745,212 |
| Words | 2,5M | 28M | 113M |
| Unique Words | 92,135 | 311,186 | 1,519,771 |
| Avg. Sen. Len. | 5 | 21 | 13 |
| Accuracy | 43% | 67% | 71% |

## 5. MT Experiments

A number of experiments were performed to evaluate various versions of our SMT systems. The experiments involved a number of steps. Processing of the corpora was performed, which included: tokenization, cleaning, factorization, conversion to lower case, splitting, and a final cleaning after splitting. Training data was then processed in order to develop the language model. Tuning was performed for each experiment. Lastly, the experiments were conducted and evaluated using a series of metrics.

Testing of the baseline system was performed using the Moses open source SMT toolkit with its Experiment Management System (EMS)7. The SRI Language Modeling Toolkit (SRILM)10 with an interpolated version of the Kneser-Key discounting (interpolate -unk discounting) was used for 5-gram language model training. We used the MGIZA++ tool for word and phrase alignment. KenLM11 was used to binarize the language model, with a lexical reordering set to use the msd-bidirectional-fe model. Reordering probabilities of phrases was conditioned on lexical values of a phrase. This considers three different orientation types on source and target phrases: monotone (M), swap (S), and discontinuous (D). The bidirectional reordering model adds probabilities of possible mutual positions of source counterparts to current and following phrases. The probability distribution to a foreign phrase is determined by "f" and to the English phrase by "e". MGIZA++ is a multi-threaded version of the well-known GIZA++ tool. The symmetrization method was set to grow-diag-final-and for word alignment processing. First, two-way direction alignments obtained from GIZA++ were intersected, so that only the alignment points that occurred in both alignments remained. In the second phase, additional alignment points existing in their union were added. The growing step adds potential alignment points of unaligned words and neighbors. The neighborhood can be set directly to left, right, top, bottom, or diagonal (grow-diag). In the final step, alignment points between words from which at least one is unaligned were added (grow-diag-final). If the grow-diag-final-and method is used, an alignment point between two unaligned words appears.

There were approximately 2 million un-tokenized Polish words contained in the TED talks, 91 million words in the OpenSubtitles corpus, and 15 million words in Europarl. Preprocessing of this training information was both automatic and manual. For each of these corpora, we trained a separate system with the settings described earlier and used it for translation of ASR output.

## 5.1. Evaluation Methods

To obtain quality measurements on the translations produced by the various SMT approaches, metrics were selected to compare the SMT translations to high quality human translations. We selected the Bilingual Evaluation Understudy (BLEU), U.S. National Institute of Standards & Technology (NIST) metric, Metric for Evaluation of Translation with Explicit Ordering (MET), and Translation Error Rate (TER) for our research.

BLEU was developed based on a premise similar to that used for speech recognition, described in 13 as: "The closer a machine translation is to a professional human translation, the better it is." So, the BLEU metric is designed to measure how close SMT output is to that of human reference translations. It is important to note that translations, SMT or human, may differ significantly in word usage, word order, and phrase length13.

To address these complexities, BLEU attempts to match variable length phrases between SMT output and reference translations. Weighted match averages are used to determine the translation score14.

The NIST metric was designed to improve BLEU by rewarding the translation of infrequently used words. This was intended to further prevent inflation of SMT evaluation scores by focusing on common words and high confidence translations. As a result, the NIST metric uses heavier weights for rarer words. The final NIST score is calculated using the arithmetic mean of the n- gram matches between SMT and reference translations. In addition, a smaller brevity penalty is used for smaller variations in phrase lengths. 15

Translation Edit Rate (TER) was designed to provide a very intuitive SMT evaluation metric, requiring less data than other techniques while avoiding the labor intensity of human evaluation. It calculates the number of edits required to make a machine translation match exactly to the closest reference translation in fluency and semantics.16,17

The Metric for Evaluation of Translation with Explicit Ordering (METEOR) is intended to take several factors that are indirect in BLEU into account more directly. Recall (the proportion of matched n-grams to total reference n-grams) is used directly in this metric. In addition, METEOR explicitly measures higher order n- grams, considers word-to-word matches, and applies arithmetic averaging for a final score. Best matches against multiple reference translations are used.18

The METEOR method uses a sophisticated and incremental word alignment technique that starts by considering exact word-to-word matches, word stem matches, and synonym matches. Alternative word order similarities are then evaluated based on those matches.

## 5.2. Translation Results

From each corpus we randomly selected texts for development and testing, 500 lines for each purpose. These lines were deleted from the corpora for more reliable evaluation. With our translation system we processed plain ASR output from randomly selected lines (PLAIN), ASR output processed with our de-normalization tool (TOOL) and ASR output processed by a human translator (HUMAN). The results are showed in Table 3.

Table 3. Results of MT.

|          | Experiment | BLEU  | NIST  | MET   | TER   |
|----------|------------|-------|-------|-------|-------|
| TED      | PLAIN      | 9.31  | 4.38  | 43.37 | 77.15 |
|          | TOOL       | 11.04 | 4.57  | 45.12 | 69.15 |
|          | HUMAN      | 16.05 | 5.34  | 49.24 | 66.42 |
| EUROPARL | PLAIN      | 31.13 | 8.14  | 68.69 | 50.69 |
|          | TOOL       | 41.24 | 8.98  | 73.16 | 46.19 |
|          | HUMAN      | 63.18 | 10.78 | 82.05 | 33.14 |
| OPENSUB  | PLAIN      | 22.78 | 4.78  | 48.12 | 69.01 |
|          | TOOL       | 32.88 | 6.69  | 52.18 | 54.17 |
|          | HUMAN      | 53.21 | 7.57  | 66.40 | 46.01 |

## 6. Future Work

This research has introduced sentence boundary detection on the text produced by the ASR systems as an area for application of the NLP technology. It is not possible to state whether the boundary detection works best for any specific data set, because the variety and amount of data. It is clear that significant accuracy was obtained on a very narrow domain data, in which many phrases were repeated, even though this data was built from rather long sentences. In our opinion, the most valuable results are those of the subtitles corpora, because their sentence length is average and the data is diverse. It must be noted that we consider accuracy results over 60% as satisfactory, because we noticed that even humans, given ASR results, are not able to reproduce original text with 100% accuracy. In the future, combining training data and using additional Polish corpora in the training phase will most likely improve the accuracy of our de-normalization system. On the basis of manual analysis of the test samples, we concluded that additional changes to and extension of the implemented rules are also required.

## 7. Conclusions

The preliminary results of the machine translations are promising. Our experiments showed that overcoming outstanding issues, such as the lack of sentence boundaries, punctuation, and capitalization in the ASR output (generally a stream of words that lack sentence boundaries, which breaks the SMT process) improves the quality of MT systems. The automatic solution is still not as good as humans, but opportunities for improvement remain.

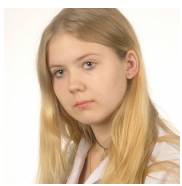

**Agnieszka Wolk**
Student and researcher at Military University of Technology in Poland. Specialist in multimedia and children game design. Certified specialist in Microsoft and Apple Technologies.